\pdfoutput=1
\documentclass[11pt]{article}
\usepackage[dvipsnames, svgnames, x11names]{xcolor}

\usepackage{ACL2023}

\usepackage{times}
\usepackage{latexsym}

\usepackage[T1]{fontenc}
\usepackage[utf8]{inputenc}

\usepackage{microtype}

\usepackage{inconsolata}

\newcommand*\samethanks[1][\value{footnote}]{\footnotemark[#1]}
\usepackage{graphicx} 
\usepackage{float} 
\usepackage{subfigure}
\usepackage{amsmath}
\usepackage{color}
\usepackage{multirow}
\usepackage{makecell}
\usepackage[frozencache=true,cachedir=minted-cache]{minted}
\usepackage{framed}
\usepackage{color}
\usepackage{courier}
\usepackage{bbding}
\usepackage{colortbl}
\definecolor{light-gray}{gray}{0.92}
\usepackage{graphicx}
\usepackage{caption}
\usepackage{subfigure}
\usepackage{enumitem}
\usepackage{booktabs}
\usepackage{framed}

\usepackage{graphicx}
\title{\textsc{CodeIE}: Large Code Generation Models are Better Few-Shot Information Extractors}

\author{Peng Li\textsuperscript{1,}\thanks{\ \ \ Equal contribution.}\quad
Tianxiang Sun\textsuperscript{2,}\samethanks\quad
Qiong Tang\textsuperscript{2}\quad
Hang Yan\textsuperscript{2}\\
{\bf Yuanbin Wu\textsuperscript{3}\quad
Xuanjing Huang\textsuperscript{2}\quad
Xipeng Qiu\textsuperscript{2,}\thanks{\ \ \ Corresponding author.}} \\
\textsuperscript{1}Academy for Engineering \& Technology, Fudan University\\
\textsuperscript{2}School of Computer Science, Fudan University \\
\textsuperscript{3}School of Computer Science and Technology, East China Normal University\\
\texttt{\{lip21,qtang22\}@m.fudan.edu.cn}, \texttt{ybwu@cs.ecnu.edu.cn} \\
\texttt{\{txsun19,hyan19,xjhuang,xpqiu\}@fudan.edu.cn}
}

\begin{document}
\maketitle
\begin{abstract}
Large language models (LLMs) pre-trained on massive corpora have demonstrated impressive few-shot learning ability on many NLP tasks. 
A common practice is to recast the task into a text-to-text format such that generative LLMs of natural language (NL-LLMs) like GPT-3 can be prompted to solve it.
However, it is non-trivial to perform information extraction (IE) tasks with NL-LLMs since the output of the IE task is usually structured and therefore is hard to be converted into plain text.
In this paper, we propose to recast the structured output in the form of code instead of natural language and utilize generative LLMs of code (Code-LLMs) such as Codex to perform IE tasks, in particular, named entity recognition and relation extraction.
In contrast to NL-LLMs, we show that Code-LLMs can be well-aligned with these IE tasks by designing code-style prompts and formulating these IE tasks as code generation tasks.
Experiment results on seven benchmarks show that our method consistently outperforms fine-tuning moderate-size pre-trained models specially designed for IE tasks (e.g., UIE) and prompting NL-LLMs under few-shot settings.
We further conduct a series of in-depth analyses to demonstrate the merits of leveraging Code-LLMs for IE tasks.\footnote{Code is available at \href{https://github.com/dasepli/CodeIE}{https://github.com/dasepli/CodeIE}}

\end{abstract}

\section{Introduction}
Information extraction (IE) aims to  recognize structured information from plain text.
It spans various tasks with diverse output structures such as named entity recognition (NER), relation extraction (RE), etc.~\cite{DBLP:conf/conll/SangM03,DBLP:journals/nle/Grishman19,DBLP:conf/acl/WangSWZLY20,DBLP:conf/naacl/ZhongC21,lu-etal-2022-unified}.
To express and address these different tasks in a unified framework, recent works propose to linearize the output structures into unstructured strings and solve the IE tasks with sequence generation models~\cite{yan-etal-2021-unified-generative, huguet-cabot-navigli-2021-rebel-relation, DBLP:conf/iclr/PaoliniAKMAASXS21, josifoski-etal-2022-genie, lu-etal-2022-unified}.
For example, given the input sentence \textit{"Steve became CEO of Apple in 1998 ."} of a NER task, 
UIE~\cite{lu-etal-2022-unified} generates the target as a sequence \texttt{"((person: Steve) (organization: Apple))"}. 

While this kind of linearizing approach achieves promising results with sufficient training data, it still performs poorly under the few-shot scenario.
For instance, compared with full-data training, the performance dropped by around 20\% when applying UIE on a 5-shot NER task CoNNL03~\cite{lu-etal-2022-unified}. 
\begin{figure}[tbp]
\vspace{6pt}
\centering  
\subfigure[Performing NER with NL-LLMs]{   
\begin{minipage}{\linewidth}
\centering    
\includegraphics[width=0.98\linewidth]{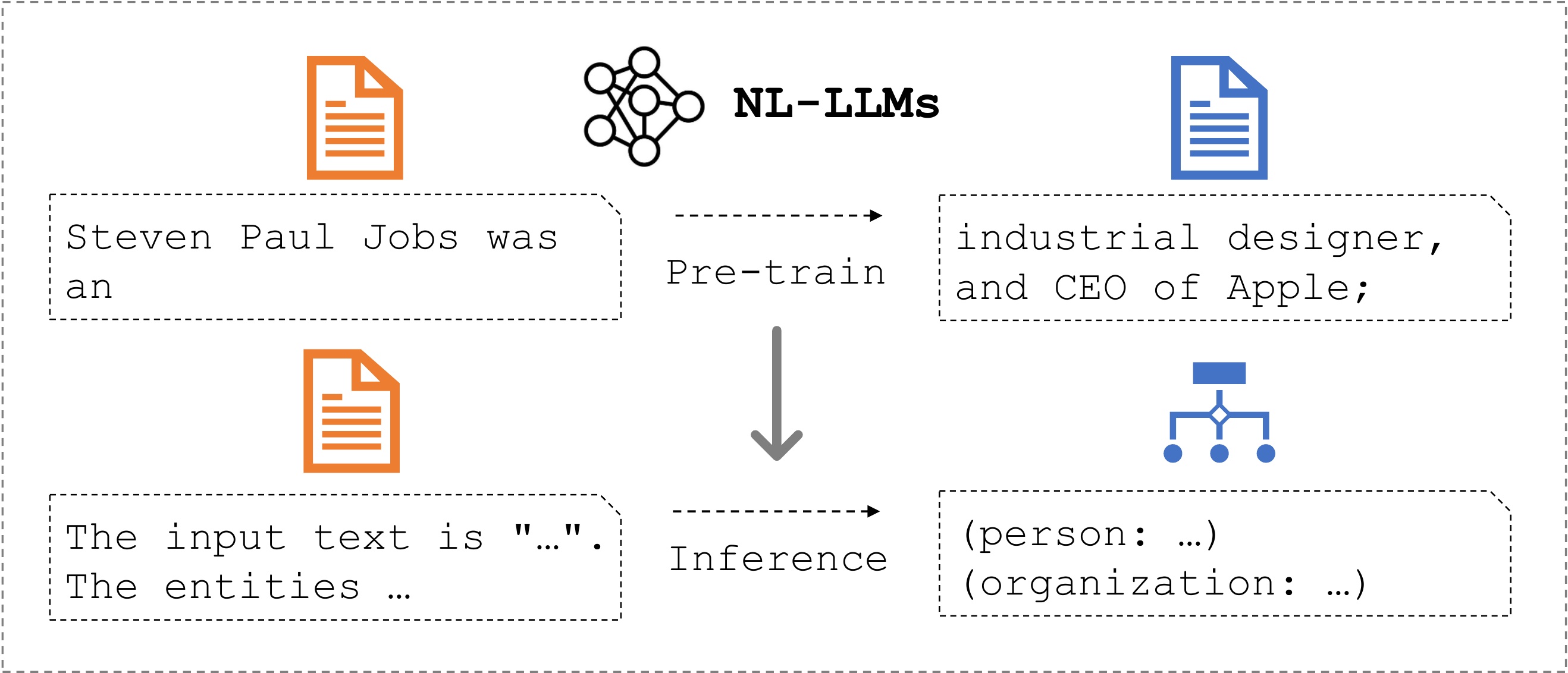}
\vspace{3pt}
\label{fig:nl-llms}  
\end{minipage}
}
\subfigure[Performing NER with Code-LLMs]{
\begin{minipage}{\linewidth}
\centering    
\includegraphics[width=0.98\linewidth]{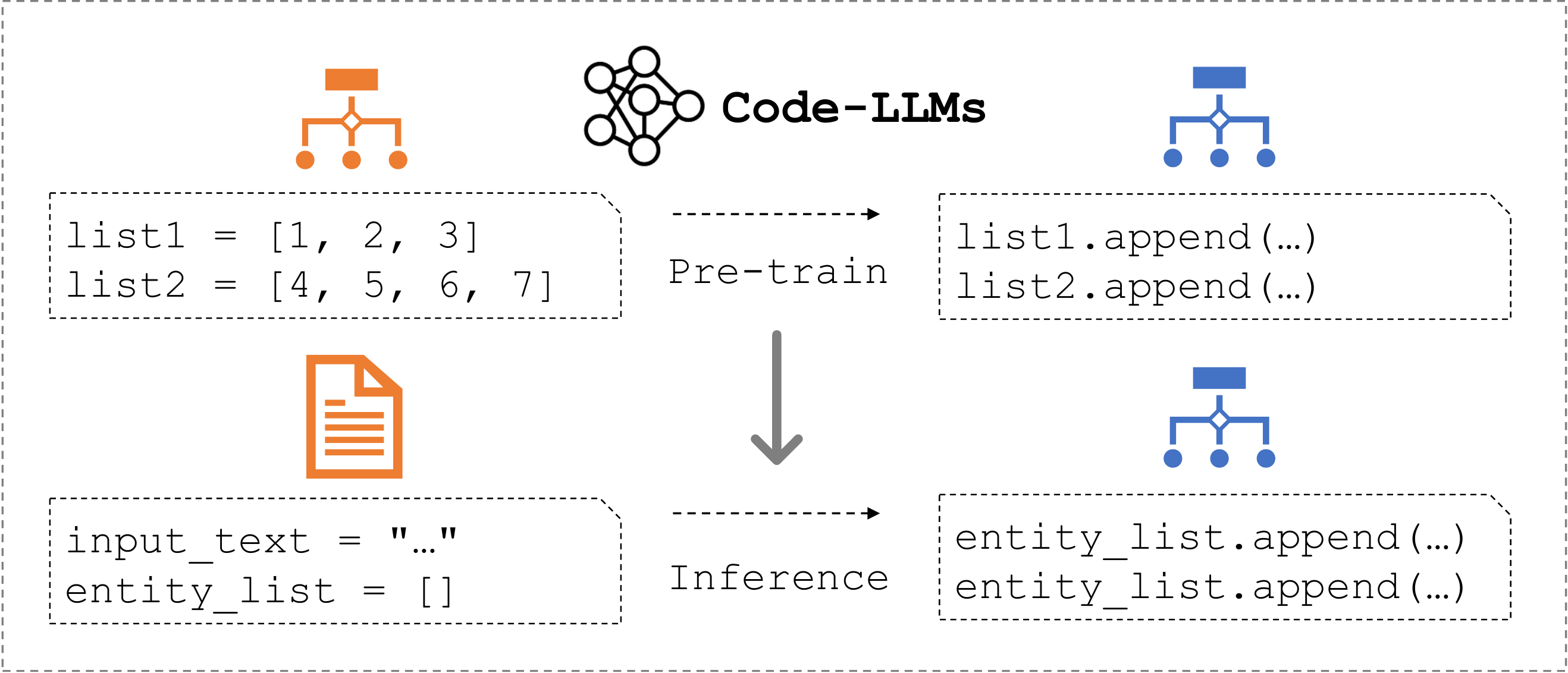}
\vspace{3pt}
\label{fig:code-elms}  
\end{minipage}
}
\vspace{-8pt}
\caption{Illustrations of performing structured NER task with NL-LLMs and Code-LLMs, respectively.
% %
% 
% %
% 
% %
% 
%
In contrast to prompting NL-LLMs with plain natural language, we utilize Code-LLMs with structured code-style prompts to mitigate the {\color[RGB]{68,114,196}output} discrepancy between the pre-training and inference stages.
}
\vspace{-10pt}
\label{fig:overview}
\end{figure}

Considering the tremendous few-shot adapting capabilities of large language models (LLMs)~\cite{DBLP:conf/nips/BrownMRSKDNSSAA20,DBLP:journals/corr/abs-2112-11446,DBLP:journals/corr/abs-2204-02311,DBLP:journals/corr/abs-2203-15556}, we manage to employ them to perform few-shot IE tasks, especially the few-shot NER task and RE task.
Typically, for NLP tasks like text classification, previous works reformulate them into text-to-text generation formats and prompt the LLMs of natural language (NL-LLMs) like GPT-3~\cite{DBLP:conf/nips/BrownMRSKDNSSAA20} to generate the answer.
In contrast, due to the complex structure inside the targets of IE tasks, linearized targets of previous works like \texttt{"((person: Steve) (organization: Apple))"} are usually "unnatural", resulting in a mismatch between the output format at the pre-training time and the inference time (see Figure \ref{fig:nl-llms}).
As a consequence, when using these flattening methods to perform IE tasks with pre-trained language models, the predicted outputs are fragile and often require complex decoding strategies to be post-processed into valid structures~\cite{lu-etal-2022-unified, josifoski-etal-2022-genie}.

In this paper, we propose to frame these two IE tasks into code generation tasks and leverage the LLMs of code (Code-LLMs) to address them.
We argue the abundant structured code information encoded in the pretrained Code-LLMs can benefit these IE tasks.
As demonstrated in Figure \ref{fig:code-elms}, it is easy to convert the text-to-structure IE task into a structure-to-structure code generation task, while transforming it into a text-to-text format can be difficult. 
Take the example input in Figure~\ref{fig:overview}, \textit{"Steve became CEO of Apple in 1998 ."}, we wrap it into a piece of Python code, and formulate the structured entity outputs as Python dictionaries with keys \texttt{"text"} and \texttt{"type"}. 
We compose them into a Python function that is semantically equivalent to the NER example, which is shown as follows:

\begin{tiny}
\begin{minted}[frame=single, baselinestretch=1.2, 
highlightcolor=light-gray!40,
highlightlines={1,2,3,4,5}]{python3}
def named_entity_recognition(input_text):
    """ extract named entities from the input_text . """
    input_text = "Steve became CEO of Apple in 1998 ."
    entity_list = []
    # extracted named entities
    entity_list.append({"text": "Steve", "type": "person"})
    entity_list.append({"text": "Apple",\ 
                        "type": "organization"})
\end{minted}
\end{tiny}
\vspace{-7pt}
\noindent After demonstrating a few training samples with the same format, we feed the code-style prompt (the highlighted lines with light grey color) into Code-LLMs and get the structured prediction.

We conduct experiments on seven benchmarks of NER and RE tasks, and carefully analyze the benefits of our approach (named \textsc{CodeIE}).
The findings are as follows:
\begin{itemize}
\setlength{\itemsep}{2pt}
\setlength{\parsep}{0pt}
\setlength{\parskip}{0pt}
  \item [1)] Prompting Code-LLMs (e.g., Codex~\cite{DBLP:journals/corr/abs-2107-03374}) with code-style inputs consistently outperforms fine-tuning UIE, a specially pre-trained model for IE tasks, and prompting NL-LLMs (e.g., GPT-3) under few-shot settings.
  \item [2)] With the same LLM (either NL-LLM or Code-LLM), the code-style prompt performs better than the linearized text prompt, demonstrating the advantage of representing structured targets with code.
  \item [3)] With the same prompt (either natural language or code), the Code-LLM (i.e., Codex) achieves better performance than the NL-LLM (i.e., GPT-3), demonstrating the merits of performing IE tasks with Code-LLMs.
  \item [4)] Compared with natural language prompts, using the code-style prompts showed higher fidelity to the output structures, i.e., the outputs have a lower structural error rate.
\end{itemize}

\begin{figure*}[tbp]
\centering  
\subfigure[Converting NER into code generation task]{
\begin{minipage}{\linewidth}
\centering    
\includegraphics[width=0.98\linewidth]{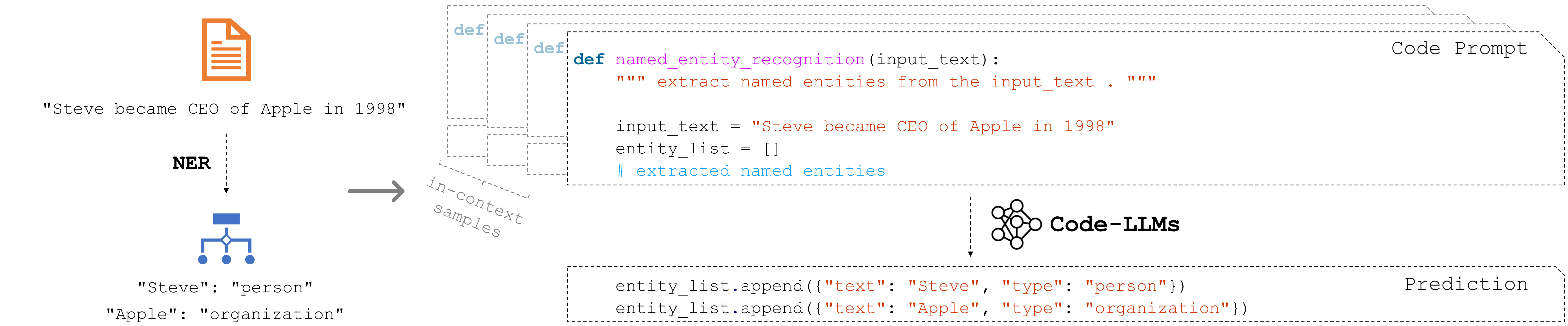}
\label{fig:codeie-ner}
\vspace{5pt}
\end{minipage}
}
\subfigure[Converting RE into code generation task]{
\begin{minipage}{\linewidth}
\centering    
\includegraphics[width=0.98\linewidth]{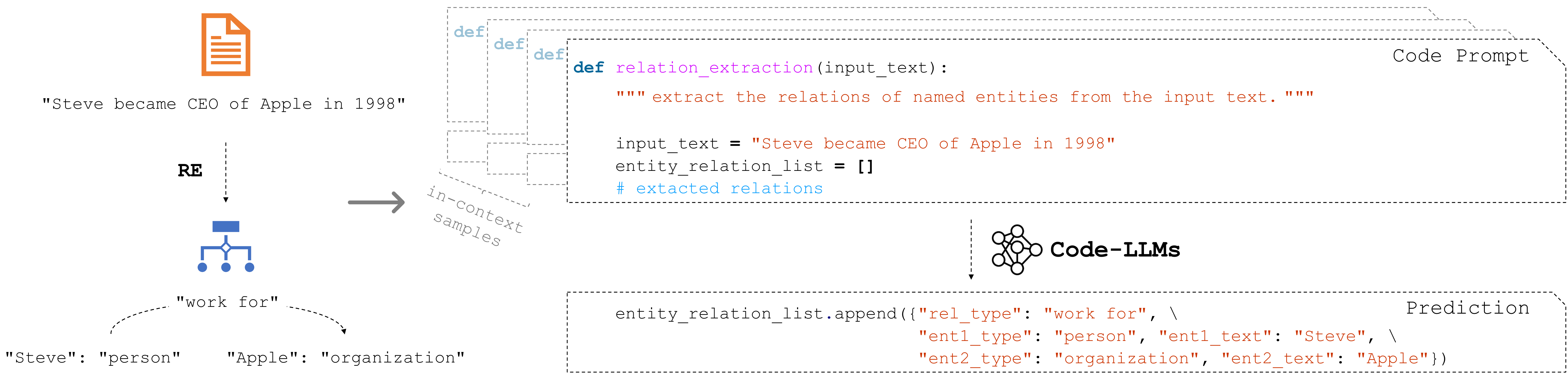}
\label{fig:codeie-re}  
% \hspace{0.25cm}
\vspace{5pt}
\end{minipage}
}
\vspace{-10pt}
\caption{The way to convert NER and RE into code generation task.}
\vspace{-5pt}
\label{fig:main}
\end{figure*}

\begin{table}
\centering
\resizebox{\linewidth}{!}{
\begin{tabular}{ccccc}
\hline
\multirow{2}{*}{\textbf{\begin{tabular}[c]{@{}c@{}}Model\\ Type\end{tabular}}} & \textbf{Generative?}                                     & \textbf{\begin{tabular}[c]{@{}c@{}}Extremely\\ Large?\end{tabular}} & \textbf{\begin{tabular}[c]{@{}c@{}}Structured\\ Pre-train?\end{tabular}} & \multirow{2}{*}{\textbf{\begin{tabular}[c]{@{}c@{}} Few-Shot\\ NER and RE\\Tasks\end{tabular}}} \\ \cline{2-4}& \begin{tabular}[c]{@{}c@{}}Unified\\ Framework\end{tabular} & \begin{tabular}[c]{@{}c@{}}Few-shot\\ Learning\end{tabular}        & \begin{tabular}[c]{@{}c@{}}Structured\\ Task\end{tabular}          &                                                                                 \\ \hline
\begin{tabular}[c]{@{}c@{}}Pre. Models\\ (e.g., UIE)\end{tabular}               &   \Checkmark                                                      & \XSolidBrush                                                                  & \Checkmark                                                                  &   {\color{red}\XSolidBrush}                                                                          \\ \hline
\begin{tabular}[c]{@{}c@{}}NL-LLMs\\ (e.g., GPT-3)\end{tabular}                 &    \Checkmark                    &        \Checkmark                                                            & \XSolidBrush                                                                  &    {\color{red}\XSolidBrush}                                                                           \\ \hline
\rowcolor{Lavender} \begin{tabular}[c]{@{}c@{}}Code-LLMs\\ (e.g., Codex)\end{tabular}              &      \Checkmark   &   \Checkmark     &    \Checkmark &   {\color{Green4}\Checkmark}                                                                              \\ \hline
\end{tabular}
} % light-gray
\caption{A high-level comparison between previous IE Models, NL-LLMs and Code-LLMs. The bottom row illustrates our approach.}
\vspace{-10pt}
\label{tab:comparison}
\end{table}

The high-level differences between previous moderate-size models, NL-LLMs, and Code-LLMs for IE tasks are summarized in Table \ref{tab:comparison}.

\section{\textsc{CodeIE}}
In this section, we first formulate the two IE tasks we focus on, named entity recognition (NER) and relation extraction (RE) in Section \ref{sec:task-formulate}. 
Then we describe how we recast these structured prediction tasks into code generation tasks (Section \ref{sec:ie2code-gen}) and prompt Code-LLMs to perform them (Section \ref{sec:icl}) under the few-shot scenario. 
We use Python language for our code generation tasks since public Python codebases are abundant and Code-LLMs are sufficiently pre-trained on them.

\subsection{Task Formulation}
\label{sec:task-formulate}
Given an input sentence $x$ with $l$ tokens $x_1, x_2,\dots,x_l$, IE tasks are to predict structured target $y$ from $x$.

The target $y$ of NER is a set of $(e, t)$ pairs, where $e$ is an entity span (e.g., \texttt{"Steve"}) and $t$ is the corresponding entity type (e.g., \texttt{"person"}).
The entity span is a sequence of tokens from $x$, and the entity type belongs to a pre-defined entity type set $T$.

The prediction target $y$ of RE is comprised of a set of triplets $(e_1, r, e_2)$, where $e_1$ and $e_2$ are two entity spans from $x$ and $r \in R$ is the semantic relation (e.g., \texttt{"work for"}) between the two entities. Here $R$ denotes a pre-defined relation type set.
In addition to extracting the relation of entities, we are often interested in also predicting the entity types $t_1$ and $t_2$ of entities $e_1$ and $e_2$ at the same time.

In the few-shot setting, we are given a small set of annotated samples $\{(x_{i}, y_{i})\}_{i=1}^{n}$ that consists of $k$ samples per class to compose a $k$-shot setting.

\subsection{Formulating IE Tasks into Code Generation Task}
\label{sec:ie2code-gen}
%% general procedure
%
In order to utilize generative Code-LLMs for IE tasks, we reformulate IE tasks as code generation tasks.
The code generation task is to predict the subsequent code sequence given an incomplete piece of code.
Hence, we can recast the input and output of the IE task into an incomplete piece of code and the code to be predicted, respectively, such that they can compose a complete piece of code that is semantically equivalent to the original sample while maintaining the syntax of the programming language.

In this work, we mainly use Python functions to represent IE tasks.
We wrap the input text $x$ into a code-style prompt $x^c$ and represent the output structure $y$ with structured Python elements, such as the list, dictionary, etc.
As shown in Figure \ref{fig:main}, for NER and RE tasks, we first transform the task name into the {\color[RGB]{216,42,254} name} of the Python function and add a {\color[RGB]{224,129,98}docstring} to illustrate the goal of the task.
% 
% %
% 
%
We assign the input text string $x$ to a variable \texttt{input\_text}.
Then we initialize an empty list to save the output and append a descriptive {\color[RGB]{9,159,254}comment} like \texttt{"\# extracted named entities"} to prompt Code-LLMs to put named entities into the list.
We pack the above code as our code prompt $x^{c}$.

For the structured target $y$, we utilize the \texttt{append} method of Python list and represent each basic information unit (e.g., a pair for NER tasks or a triplet for RE tasks) as a {\color[RGB]{224,129,98}Python dictionary}.
Hence, the target $y^c$ to be predicted by Code-LLMs is reformulated into a list of dictionaries.
For NER, we add Python dictionaries with keys \texttt{"text"} and \texttt{"type"}
to the list, where the values of the dictionaries are the corresponding entity span and entity type.
For RE, we similarly add dictionaries with keys \texttt{"rel\_type"}, \texttt{"ent1\_type"}, \texttt{"ent1\_text"}, \texttt{"ent2\_type"}, and \texttt{"ent2\_text"} to the list to represent the structured target.

The Code-LLM is expected to complete the list conditioning on the function name, docstring, and input text. Figure~\ref{fig:main} shows examples of formulating an original IE sample into a code-style one.

It is worth noting that the design space of the code-style prompt is large and hard to be fully explored. The formulation described above is a straightforward instance using Python. We also explore several other formulations to recast IE tasks into code generation tasks, which can be found in Appendix~\ref{sec:code-prompts}. 

\subsection{Prompting Code-LLMs with In-Context Demonstrations}
\label{sec:icl}
Despite the carefully designed prompt, it is non-trivial to perform IE tasks by prompting Code-LLMs without any samples. Therefore, it is necessary to let Code-LLMs be aware of a few labeled samples in typical few-shot settings.

With the increasing size of pre-trained language models, fine-tuning is becoming more and more expensive or even infeasible since recent LLMs are usually released as black-box APIs~\cite{DBLP:conf/icml/SunSQHQ22}. 
Hence, instead of fine-tuning Code-LLMs on the few-shot dataset, we explore including the labeled samples in the context and performing in-context learning~\cite{DBLP:conf/nips/BrownMRSKDNSSAA20}. 
We select $n$ samples $\{(x_{i}, y_{i})\}_{i=1}^{n}$ from the training dataset and convert them to corresponding code-style pairs $\{(x_{i}^c, y_{i}^c)\}_{i=1}^{n}$.
We concatenate them as a string to compose the in-context demonstrations $x_{1}^c \oplus y_{1}^c \dots x_{n}^c \oplus y_{n}^c $.
Given an arrived test sample $x$, we first convert it to a code prompt $x^c$ and prepend the demonstration context, i.e., $x_{1}^c \oplus y_{1}^c \dots x_{n}^c \oplus y_{n}^c\oplus x^c$. 
After feeding the constructed input into the Code-LLM, we are expected to get an output $y^c$ that is formatted as the same as $y_1^c$, $y_2^c$, $\dots y_n^c$ (see Figure \ref{fig:main}).
We find that $y^c$ almost always retains the syntax of Python language and is easy to be converted back to its original structure $y$.

\section{Experiments}
\subsection{Setup}
\paragraph{Datasets} We evaluate our approach on NER task with CoNLL03~\cite{DBLP:conf/conll/SangM03}, ACE04~\cite{DBLP:conf/lrec/DoddingtonMPRSW04} and ACE05-E\cite{ace05}. For relation extraction, we evaluate on datasets CoNLL04~\cite{DBLP:conf/conll/RothY04}, ACE05-R~\cite{ace05}, NYT~\cite{DBLP:conf/pkdd/RiedelYM10} and SciERC~\cite{luan-etal-2018-multi}. 
Table \ref{tab:dataset_stats} shows the dataset statistics.
We follow ~\citet{lu-etal-2022-unified} to preprocess all these datasets.

\begin{table}
\centering
\resizebox{\linewidth}{!}{
\begin{tabular}{l|ll|lll}
\toprule
          & |Ents| & |Rels| & \#Train & \#Val & \#Test \\ \midrule
CoNLL03   & 4     & -     & 14,041  & 3,250 & 3,453  \\
ACE04     & 7     & -     & 6,202   & 745   & 812    \\
ACE05-E & 7     & -     & 7299    & 971   & 1060   \\
CoNLL04   & 4     & 5     & 922     & 231   & 288    \\
ACE05-R & 7     & 6     & 10,051  & 2,420 & 2,050  \\
NYT       & 3     & 24    & 56,196  & 5,000 & 5,000  \\
SciERC    & 6     & 7     & 1,861   & 275   & 551    \\ \bottomrule
\end{tabular}
}
\caption{Statistics of the datasets used in our experiments. |Ents| and |Rels| denote the number of entity types and relation types. \#Train, \#Val and \#Test denote the sample number in each split.}
\vspace{-5pt}
\label{tab:dataset_stats}
\end{table}

\begin{table*}[htb]
\resizebox{\linewidth}{!}{
\begin{tabular}{lccccccccc}
\toprule
\multicolumn{1}{c}{}                        &                                                                         & \multicolumn{3}{c}{\textbf{Entity}}                    & \multicolumn{4}{c}{\textbf{Relation}}                                                            &                                \\ 
\cmidrule(l{3pt}r{3pt}){3-5}% \cline{3-9}
\cmidrule(l{3pt}r{3pt}){6-9}
\multicolumn{1}{c}{\multirow{-2}{*}{\textbf{Model}}} & \multirow{-2}{*}{\begin{tabular}[c]{@{}c@{}}\textbf{Prompt}\\ \textbf{Type}\end{tabular}} & \textbf{CoNLL03} & \textbf{ACE04} & \textbf{ACE05-E} & \textbf{CoNLL04}& \textbf{ACE05-R} & \textbf{NYT}                           & \textbf{SciERC} & \multirow{-2}{*}{\textbf{AVG}} \\ 
\midrule
\multicolumn{10}{c}{\textbf{Full Data}} \\ \midrule
Pre. SoTA                                    & -  & \textbf{93.21} & 86.84          & 84.74               & 73.60   & 65.60                         & 92.70                                  & 35.60           & 76.04                          \\
UIE-large                                  & text   & 92.99   & \textbf{86.89} & \textbf{85.78}      & \textbf{75.00}          & \textbf{66.06}        & - & \textbf{36.53}  & -                \\ 
\midrule
\multicolumn{10}{c}{\textbf{Few Shot}}   \\
\midrule
\multicolumn{2}{c}{ \#shot (\#sample)}
   & 5 (25)  & 2 (16)         & 2 (16)                   & 5 (25)       & 2 (14)                       & 1 (24)                                 & 2 (16)          & \multicolumn{1}{l}{}           \\ 
  \midrule
T5-base   & text  & $\text{33.68}_{\pm 29.17}$ &   $\text{7.25}_{\pm 12.00}$ &   $\text{9.09}_{\pm 15.74}$ &    $\text{14.56}_{\pm 13.87}$      &    $\text{0.00}_{\pm 0.00}$       &         $\text{5.59}_{\pm 9.68}$             &        $\text{0.00}_{\pm 0.00}$         &   $\text{10.02}$                             \\
UIE-base     & text   & 
$\text{70.37}_{\pm 0.54}$         & $\text{44.31}_{\pm 1.61}$              &         $\text{39.71}_{\pm 0.91}$       &  
$\text{45.63}_{\pm 1.50}$  &
$\text{8.69}_{\pm 1.41}$ &                      -            &  $\text{5.69}_{\pm 0.49}$  & -                               \\
T5-large                                   & text   &            $\text{53.08}_{\pm 7.71}$ &  $\text{24.67}_{\pm 5.26}$  &          $\text{24.31}_{\pm 4.74}$ 
&             $\text{10.03}_{\pm 8.75}$&           $\text{1.41}_{\pm 0.74}$              &                  $\text{15.29}_{\pm 8.76}$                      &     $\text{0.25}_{\pm 0.43}$            &                $18.43$                \\
UIE-large                                   & text  &      $\textbf{70.62}_{\pm 3.22}$   &   $\textbf{45.08}_{\pm 3.63}$             &   $\textbf{43.03}_{\pm 2.26}$   
&        $\textbf{47.68}_{\pm 2.29}$&       $\textbf{9.59}_{\pm 4.89}$                       &                                        - & $\textbf{7.30}_{\pm 2.01}$                &               -                 \\ \midrule
GPT-3  & text   &   $\text{68.84}_{\pm 1.29}$ &   $\text{45.51}_{\pm 0.23}$  &   $\text{48.93}_{\pm 0.49}$    &  $\text{39.67}_{\pm 2.44}$         &   $\text{5.13}_{\pm 1.24}$       &        $\text{16.07}_{\pm 4.67}$                     &       $\text{4.39}_{\pm 0.98}$  &     $\text{32.65}$                           \\
GPT-3  & code                                         & $\text{81.00}_{\pm 1.49}$                &                 $\text{53.44}_{\pm 1.44}$   &       $\text{52.98}_{\pm 1.53}$         &    $\text{51.33}_{\pm 1.34}$               &  $\text{12.33}_{\pm 2.06}$      & $\text{24.81}_{\pm 1.90}$      &  $\text{4.67}_{\pm 0.67}$    &    $\text{40.08}$                            \\
Codex  & text    & 
$\text{72.66}_{\pm 0.66}$ &
$\text{49.58}_{\pm 1.37}$        & $\text{49.55}_{\pm 1.14}$          &       $\text{47.30}_{\pm 2.25}$     &    $\text{10.08}_{\pm 2.06}$        &    $\text{24.63}_{\pm 6.74}$       & $\text{5.40}_{\pm 2.65}$  &           
$\text{37.03}$               \\
\rowcolor{Lavender} Codex                                       & code     & $\textbf{82.32}_{\pm 0.37}$          & $\textbf{55.29}_{\pm 0.37}$     &  $\textbf{54.82}_{\pm 2.09}$  
&  $\textbf{53.10}_{\pm 2.02}$    & $\textbf{14.02}_{\pm 3.27}$   & $\textbf{32.17}_{\pm 1.46}$    &    $\textbf{7.74}_{\pm 1.54}$        & 
$\textbf{42.78}$                \\ 
\bottomrule
\end{tabular}
}
\caption{Experiment performances on NER and RE benchmarks. Our approach is highlighted with light grey. The full data fine-tuning performances come from UIE. For the few-shot setting, we evaluate T5-base, UIE-base, T5-large and UIE-large with fine-tuning, and GPT-3 and Codex by few-shot prompting with two different prompt types. 
The text prompt is the structured extraction language (SEL) introduced by UIE.
The code format is introduced in Section \ref{sec:ie2code-gen}.
We set the shot number (\#shot) and the corresponding sample number (\#sample) differently to fit into the GPT-3 maximum length limitation (4097 tokens).}
\label{tab:main-results}
\end{table*}

\paragraph{Code-LLMs} 
For Code-LLMs, we conduct experiments mainly with the \texttt{code-davinci-002} version Codex from OpenAI.
Codex is a large language model adapted from GPT-3 and further pre-trained on open-source codebases. 
The \texttt{code-davinci-002} version Codex supports 8k input tokens at most.
We get the model predictions by querying OpenAI API\footnote{https://openai.com/api} in the few-shot in-context prompting way. We generate up to 280 tokens with greedy decoding.

\paragraph{Baselines} We compare our approach with two kinds of few-shot learning methods:
\begin{itemize}
\setlength{\itemsep}{2pt}
\setlength{\parsep}{0pt}
\setlength{\parskip}{0pt}
  \item [1)] \textbf{Fine-tuning} We fine-tune the base and large versions of two moderate-size pre-trained models: T5 and UIE.
  T5 is a sequence-to-sequence model pre-trained on large-scale text corpora.
  UIE is a model further pre-trained from T5 on the structured datasets.
  UIE utilizes the textual structured extraction language (SEL) to express the output structures.
  We use the same approach and parameters with~\citet{lu-etal-2022-unified} when fine-tuning T5 and UIE.
  
  \item [2)] \textbf{Prompting} We compare our approach with prompting NL-LLMs, in particular GPT-3.
  We mainly experiment with the \texttt{text-davinci-002}.
  We use a text prompt, of which the format is slightly modified from SEL.
  As shown in Figure \ref{fig:nl-llms}, given an input text $x$, the text prompt and output format are like \texttt{"The text is $x$. The named entities in the text: "} and \texttt{"((person: ...)(organization:...))"}, respectively.
  See Appendix \ref{sec:text-prompts} for more details of the text prompt.
  The approach and super-parameters of NL-LLMs prompting and Code-LLMs prompting are identical.
\end{itemize}

\paragraph{Few-Shot Setting} For each IE task, we randomly sample $k$ training samples for each entity or relation type to construct a $k$-shot training set.
The value of $k$ varies across different datasets to satisfy the maximum length limitation (4097) of GPT-3.
To be compatible with datasets that contain samples with empty targets, we regard those empty-target samples as an additional class and include $k$ samples belonging to that class in the training set. 

\paragraph{Evaluation} Same as previous work~\cite{lu-etal-2022-unified},
we use \textbf{Entity F1} and \textbf{Relation Strict F1} as the evaluation metrics for NER tasks and RE tasks, respectively. 
Under these metrics, an entity span prediction is correct if its offsets and entity type match the golden entity.
And a relation prediction is correct if the relation type is correct and the corresponding offsets and types of its entities are correct.
Since few-shot training is of high variance, we perform 3 runs with different random seeds for each experiment and report the mean and standard deviation of the metric.

\subsection{Results}
\paragraph{LLMs vs. Moderate-sized Models}
As shown in Table \ref{tab:main-results}, 
LLMs (GPT-3 and Codex) achieve superior performance over moderate-sized models (T5 and UIE) under few-shot settings, demonstrating a strong few-show learning ability on IE tasks. Especially, on average performance over the seven considered benchmarks, our proposed \textsc{CodeIE} (Codex $+$ code prompt) achieves the best results, improving T5-large and T5-base by 132\% and 327\%, respectively.
In addition, under 1-shot learning settings, \textsc{CodeIE} improves the performance of UIE-large by more than 60\% on CoNLL03 and CoNLL04 benchmarks (see Table \ref{tab:1-shot} in the Appendix).

\begin{figure}[tbp]
\centering
\includegraphics[width=0.95\linewidth]{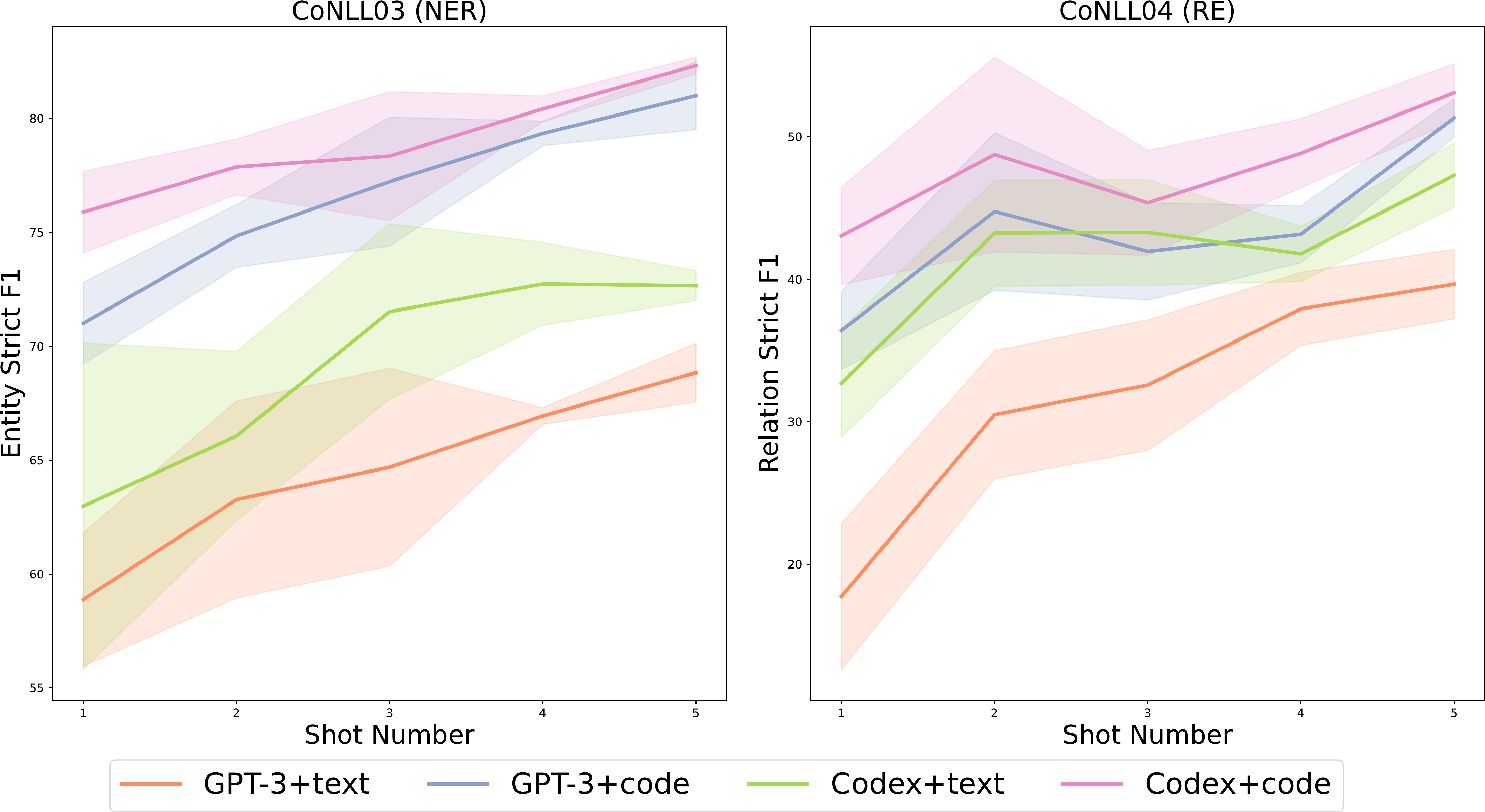}
\caption{Performance with different shot numbers on CoNLL03 (NER) and CoNLL04 (RE) datasets.}
\vspace{-8pt}
\label{fig:shot_num}
\end{figure}

\paragraph{Code Prompt vs. Text Prompt}
We then compare the performance of code prompt vs. text prompt when using the same LLM, i.e., comparing $\langle$GPT-3 $+$ text prompt$\rangle$ with $\langle$GPT-3 $+$ code prompt$\rangle$ and comparing $\langle$Codex $+$ text prompt] with $\langle$Codex $+$ code prompt$\rangle$. As a result, we find that prompting LLMs with code yields significant improvement (23\% for GPT-3 and 16\% for Codex). What is surprising is that code prompt is even more beneficial to GPT-3, which is not specifically trained on code data.

\paragraph{Code-LLMs vs. NL-LLMs}
When using the same kind of prompt and comparing the used LLMs, i.e., comparing $\langle$GPT-3 $+$ text prompt$\rangle$ and $\langle$Codex $+$ text prompt$\rangle$ and comparing $\langle$GPT-3 $+$ code prompt$\rangle$ and $\langle$Codex $+$ code prompt$\rangle$, we find that Codex consistently surpasses GPT-3, demonstrating that code pre-training can be beneficial to IE tasks.

\paragraph{Different Shot Numbers}
We further compare these approaches under different shot numbers on CoNLL03 and CoNLL04.
As shown in Figure \ref{fig:shot_num}, we can see that the obtained phenomenons still hold when increasing the number of shots.

\begin{table}
\centering
\resizebox{0.9\linewidth }{!}{
\begin{tabular}{clcc}
\toprule
\textbf{Model}                  & \multicolumn{1}{c}{\textbf{\begin{tabular}[c]{@{}c@{}}Prompt\\ Design\end{tabular}}} & \textbf{\begin{tabular}[c]{@{}c@{}}Entity\\ CoNLL03\end{tabular}} & \textbf{\begin{tabular}[c]{@{}c@{}}Relation\\ CoNLL04\end{tabular}} \\ 
\midrule
\multirow{2}{*}{GPT-3}  & \cellcolor{Lavender}struct lang 
& \cellcolor{Lavender}$\text{68.84}_{\pm 1.29}$ 
&  \cellcolor{Lavender}$\text{39.67}_{\pm 2.44}$                                           \\
 & natural lang&  $\text{46.36}_{\pm 12.56}$ 
&  $\text{40.90}_{\pm 3.67}$                                           \\ \midrule
\multirow{4}{*}{Codex} & \cellcolor{Lavender}func def
&   \cellcolor{Lavender}$\text{82.32}_{\pm 0.37}$                  &    \cellcolor{Lavender}$\text{53.10}_{\pm 2.02}$                                         \\ & class init  & $\text{81.29}_{\pm 0.72}$                  &    $\text{52.32}_{\pm 0.94}$      \\  & func exec      &$\text{84.05}_{\pm 1.24}$                  &    $\text{53.32}_{\pm 3.47}$  \\ & func init-   & $\text{81.95}_{\pm 1.01}$  & $\text{53.59}_{\pm 1.10}$  \\ \hline
\end{tabular}
}
\caption{Performance of different prompt designs. "struct lang" and "func def" are the "text" and "code" prompt types respectively in our main experiments.}
\label{tab:prompt-design}
\end{table}

\paragraph{Different Prompt Designs}
The design of the prompt can be an important factor affecting the model performance~\cite{DBLP:journals/corr/abs-2202-12837}.
Hence, we explore additional prompt designs for both text prompt and code prompt. The detailed prompt deigns can be found in Appendix \ref{sec:prompt-design}.
The experimental results are shown in Table~\ref{tab:prompt-design}, from which we find that code prompts consistently outperform text prompts. Hence, the superior performance of using code prompts is mainly contributed by the code style instead of some specific instance of prompt design.

\paragraph{Different LLMs}
To verify the versatility of the proposed approach and the observed findings, we further conduct experiments with \texttt{text-davinci-001} version of GPT-3 and \texttt{code-davinci-001} version of Codex. 
As shown in Table \ref{tab:diff-elms}, the previous findings still hold across the two different versions.

\begin{figure}[tbp]
\centering
\includegraphics[width=0.9\linewidth]{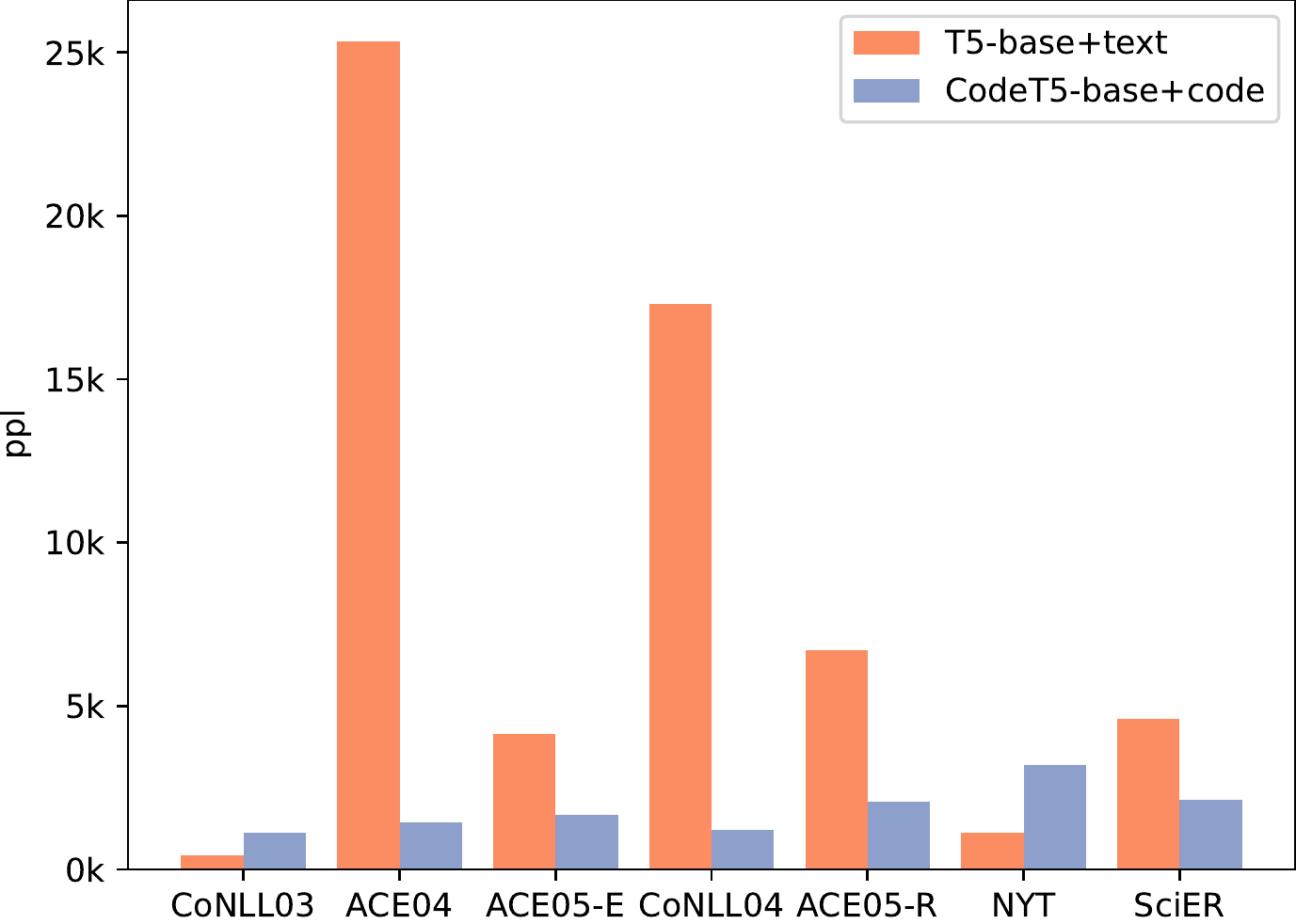}
\caption{Format consistency between the input format and the model (measured by perplexity) for text prompt and code prompt on 7 datasets.}
\label{fig:format-consist-base}
\end{figure}

\section{Analysis}
\label{sec:analysis}
To take a closer look at the difference between prompting NL-LLMs with textual format input and prompting Code-LLMs with code format input, in this section, we define several informative metrics and conduct in-depth analyses to shed some light on the following question: \textit{what contributes to the final performance of \textsc{CodeIE} for IE tasks?} % todo: nonsense

\subsection{Format Consistency}
We can see from Figure \ref{fig:nl-llms} that an apparent inappropriateness to use NL-LLMs for IE tasks is the inconsistency between the structured output format at inference time and NL-LLMs that are trained on natural language at pre-training time, while the format of code-style output aligns well with Code-LLMs.
It has been evidenced that adapting pre-trained models to downstream tasks in a manner that is well aligned with it pre-training paradigm usually achieves better few-shot learning performance. Hence we assume \textit{the promising performance of \textsc{CodeIE} partly comes from the better format consistency between the code-style sample and the pretrained code model}.

To verify this hypothesis, given a sample, we compare the perplexities of a pre-trained language model on its text format and a pre-trained code model on its code format.
Formally, given a generative model $M$, the conditional perplexity $ppl$ of a sample $(x,y)$ is as follows,
\begin{align}
    ppl_{M}(y|x)
    &= \prod_{i=1}^mP_{M}(y_i|y_1\cdots y_{i-1},x)^{-\frac{1}{l}}.
\end{align}
For an original IE sample $(x,y)$, we first transform it to its natural language text pair $(x^{t}, y^{t})$ and its code piece pair $(x^c, y^c)$, and then compute the conditional perplexity of them with the language model $M^{nl}$ and the code model $M^{c}$, respectively, i.e., the $ppl_{M^{nl}}(y^{t}|x^{t})$ and $ppl_{M^{c}}(y^{c}|x^{c})$.
A lower conditional perplexity means the output format aligns well with the pre-training distribution of the model.

Since LLMs usually limit user access by their black-box APIs, we instead utilize two agent models T5 ~\cite{DBLP:journals/jmlr/RaffelSRLNMZLL20} and CodeT5 ~\cite{wang-etal-2021-codet5} to calculate the perplexities. 
CodeT5 is a variant of T5 model that is further pre-trained on code data.
We calculate the perplexities on the previous seven datasets with the base verison of the two models, namely T5-base and CodeT5-base.
Figure \ref{fig:format-consist-base} shows the mean perplexities of two base version models on the training samples of each task.
We can observe the perplexity of the text format outputs measured by T5-base is usually larger than code format outputs measured by CodeT5-base. 
That means, transforming IE samples to code formats can better align with the data distribution of code pre-training and therefore the pre-trained code language model.

\begin{figure}[tbp]
\centering
\includegraphics[width=0.9\linewidth]{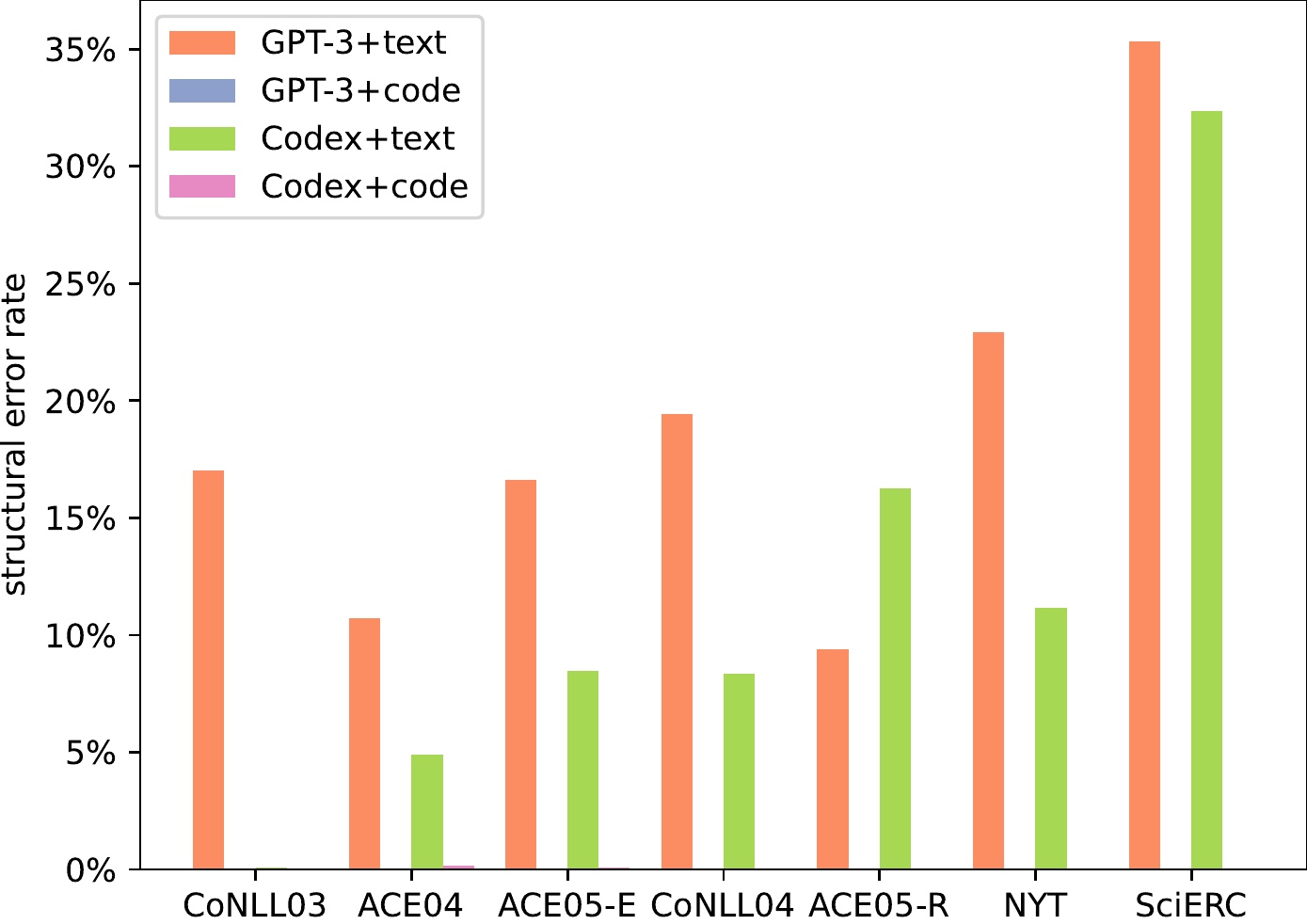}
\caption{Structural error rate of different combinations of LLM and prompting methods. Prompting LLMs with code exhibits higher structure fidelity.}
\label{fig:model-faithful}
\end{figure}

\subsection{Model Fidelity}
Besides the low format consistency of prompting ML-LLMs, we find that NL-LLMs are more likely to generate outputs with structural and semantic errors when performing few-shot IE tasks than Code-LLMs.
In other words, \textit{Code-LLMs seem to be more faithful to the demonstrated few-shot samples than NL-LLMs}.
To quantitatively measure the model fidelity, we define two metrics:
\paragraph{Structure Fidelity}
Structure fidelity measures how faithful the model is to the structure of demonstrations provided in the context.
This can be simply measured by calculating the structural error rate, which is the proportion of generated samples with structural errors.
% todo: parser?
In particular, we construct a parser with a series of hand-written rules to transform the model-generated outputs back to the desired format and filter out samples with invalid structures.
Figure \ref{fig:model-faithful} demonstrates the structure fidelity of different models with different prompts on the seven benchmarks.
Results show that the outputs generated by GPT-3 and Codex using text prompts are fragile while using code prompts tends to generate nearly zero structurally erroneous samples.
Besides, with the same text prompt, Codex tends to generate fewer structurally errant samples than GPT-3, demonstrating its superior understanding ability on general structured input instead of being limited to existing programming languages.

\begin{table}
\centering
\resizebox{0.92\linewidth }{!}{
\begin{tabular}{ccc}
\toprule
\textbf{Task} & \textbf{Error Type}                                                 & \textbf{Samples}                                                                                                                          \\ \midrule
NER           & \begin{tabular}[c]{@{}c@{}}Entity type\\ not in $T$\end{tabular}   & \begin{tabular}[c]{@{}c@{}}currency, company, time, event,\\ profession,organizational indicator,\\ finanical, object, event\end{tabular} \\ \midrule
RE            & \begin{tabular}[c]{@{}c@{}}Relation type\\ not in $R$\end{tabular} & \begin{tabular}[c]{@{}c@{}}called, organization, person, relate,\\ specialize, assumption, cause, assign\end{tabular}                     \\ \bottomrule
\end{tabular}
}
\caption{Semantically errant samples detected in our experiments. These errant samples mainly came from GPT-3.}
\label{tab:error-samples}
\end{table}

\begin{figure*}[tbp]
\centering
\includegraphics[width=0.75\linewidth]{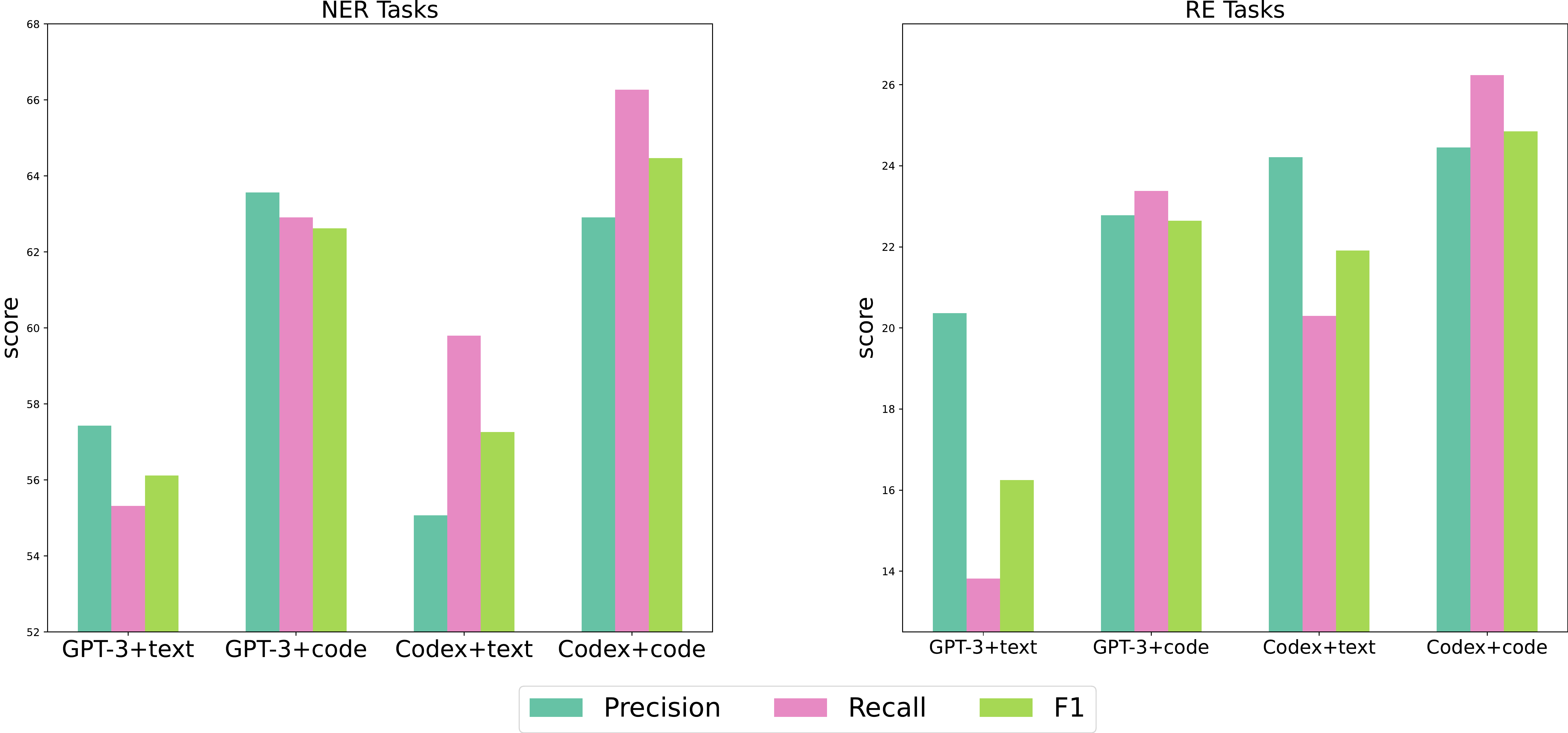}
\caption{Model Performance Details on NER and RE Tasks. We report the averaged metric scores of all the NER or RE datasets.}
\label{fig:perform-details}
\end{figure*}

\paragraph{Semantic Fidelity}
Another measurement of model fidelity is semantic fidelity, which is designed for those samples that have a valid structure and can succeed in our parser but are semantically incorrect.
The difference between the defined semantic fidelity and the conventional prediction error is that semantic fidelity mainly considers model behaviours that violate the formulation of the task, e.g., predicting an entity type that does not exist in the given entity type set or extracting an entity span that does not appear in the input text.
Some example semantic errors detected in our experiments are listed in Table \ref{tab:error-samples}.
We report the statistical result of the tasks in Table \ref{tab:ner-error} and Table \ref{tab:re-error} in the Appendix.
As a result, we find that GPT-3 generated more semantic errors than Codex though some of the errors seem to be "correct" but are out of the pre-defined class set.
In a nutshell, GPT-3 tends to generate free-form results and Codex is more faithful to the demonstrations provided in the context and therefore is more predictable for IE tasks.

\subsection{Fine-grained Performance}
In addition, we conduct a fine-grained evaluation to compare different approaches.
In addition to the F1 score, precision and recall are also important metrics for NER and RE tasks.
To investigate how different LLMs and prompting methods affect precision and recall, we report the two metrics in Figure~\ref{fig:perform-details}.
Results show that: (a) The code prompt improves model performance in both precision and recall;
(b) Compared with GPT-3, Codex achieves higher recall and comparable precision on NER tasks and and achieves both higher precision and recall on RE tasks.

\section{Related Work}
% information extraction

\paragraph{Generative Information Extraction}

Generative information extraction which frames IE tasks as token generation tasks receive more attention recently due to their potential to unify different tasks~\cite{yan-etal-2021-unified,josifoski-etal-2022-genie}.
\citet{yan-etal-2021-unified} designs various ways to linearize entities into a sentence to unify various named entity recognition subtasks.
TANL~\cite{DBLP:conf/iclr/PaoliniAKMAASXS21} uses augmented language to improve the effect of generative models.
\citet{lu-etal-2022-unified} also proposes a structured extraction language (SEL) and pre-trains their UIE model with this language on multiple structured datasets.
These works linearize the structure output of IE tasks into text format to align the pre-trained models.
Different from them, we propose to recast the structural samples of IE tasks into structural code format and utilize aligned pre-trained code models to perform the tasks.

\paragraph{Code-LLMs for Complex Tasks}
Recent works show Code-LLMs perform better on complex tasks like commonsense and symbolic reasoning \cite{DBLP:journals/corr/abs-2210-07128,DBLP:journals/corr/abs-2210-02875}, mathematical logic~\cite{DBLP:journals/corr/abs-2210-09261} and  event argument prediction~\cite{DBLP:journals/corr/abs-2210-12810} tasks.
We focus on the two mainstream IE tasks different from them, i.e., NER and RE.
Besides, in-depth analyses are conducted to provide more insights.

\paragraph{LLMs for Few-Shot NER and RE}
While LLMs like GPT-3 have shown strong few-shot learning abilities in many NLP tasks, limited works have explored their capabilities on typical IE tasks like NER and RE.
\citet{DBLP:journals/corr/abs-2108-11857} evaluate GPT-2~\cite{radford2019language} on open-domain NER tasks with few-shot demonstrating.
A recent work~\cite{DBLP:journals/corr/abs-2203-08410} tests the performance of GPT-3 on biomedical NER and RE tasks and finds it underperforms compared to fine-tuning smaller pretrained models.
Its concurrent work~\cite{DBLP:conf/emnlp/AgrawalHLKS22} finds that GPT-3 performs well on few-shot clinical IE tasks.
We conduct our experiments on more general NER and RE datasets and find GPT-3 can achieve comparable performance to fine-tuning the UIE model.
Besides, we successfully employ the LLMs of code with better performances for these IE tasks.

\section{Conclusion}
We propose the first work to utilize the structured Code-LLMs with code-style prompts to perform the few-shot NER and RE tasks.
Experiments show our approach consistently surpasses the UIE models and the NL-LLMs counterpart under the few-shot setting.
We conducted extensive analysis and find the performances come from better format consistency and model fidelity, etc.
We think these analyzes can facilitate future work.
As the further works, we will employ \textsc{CodeIE} on more IE tasks in different domains, and inspect the robustness of it.

\section*{Limitations}
Though our approach demonstrates better performances than the baseline models, how to design a good code-format prompt has not been fully inspected.
Besides, we mainly conduct experiments on the black-box GPT-3 and Codex models but they are not open-sourced and querying the GPT-3 model cost the economic budget.
And the use of LLMs may bring environmental pollution.
Another limitation of our approach is that the Code-LLMs mainly trained on programming language datasets with English annotations.
Exploring our model on non-English datasets (like Chinese datasets) is the future work.

\section*{Acknowledgements}
We would like to express our gratitude to the reviewers for their helpful comments and suggestions. We are also very grateful to Yaojie Lu for his friendly assistance during our experiments.
This work was supported by the National Natural Science Foundation of China (No. 62236004 and No. 62022027) and CCF-Baidu Open Fund.

\bibliography{anthology,custom}
\bibliographystyle{acl_natbib}

\appendix

% \section{Structure Linearization Methods} \label{sec:appendix}

\section{Prompt Format Design} 
\label{sec:prompt-design}
\subsection{Code Format Prompts}
\label{sec:code-prompts}
We design several code-style prompt formats.
We use the input sentence \textit{"Steve became CEO of Apple in 1998 ."} and the corresponding entities ("Steve": person, "Apple": organization) and relations ("work for" of the two entities "Steve" and "Apple") as a running sample for the NER and RE tasks.

The name of the format design is denoted with different \texttt{font}. we demonstrate the Python format prompt for NER and RE tasks. The prompt part is highlighted with \colorbox{light-gray}{grey color}
and the following codes are the expected output.
We list the designed format as follows:\\

% \newpage
\noindent\texttt{1. func def:} our main code format prompt to transform the IE tasks into code formats.\\

\noindent For the NER task, the format is
\begin{tiny}
\begin{minted}[frame=single, baselinestretch=1.2, 
highlightcolor=light-gray!40,
breaklines,
highlightlines={1,2,3,4,5}]{python3}
def named_entity_recognition(input_text):
    """ extract named entities from the input_text . """
    input_text = "Steve became CEO of Apple in 1998 ."
    entity_list = []
    # extracted named entities
    entity_list.append({"text": "Steve", "type": "person"})
    entity_list.append({"text": "Apple","type": "organization"})
\end{minted}
\end{tiny}

\noindent For the RE task, the format is
\begin{tiny}
\begin{minted}[frame=single, baselinestretch=1.2, 
highlightcolor=light-gray!40,
breaklines,
highlightlines={1,2,3,4,5}]{python3}
def relation_extraction(input_text):
    """ extract the relations of named entities from the input_text . """
    input_text = "Steve became CEO of Apple in 1998"
    entity_relation_list = []
    # extacted relations
    entity_relation_list.append({"rel_type": "work for", "ent1_type": "person", "ent1_text": "Steve", "ent2_type": "organization", "ent2_text": "Apple"})
\end{minted}
\end{tiny}

\noindent\texttt{2. class init}: we describe the IE tasks with the Python class.\\

\noindent For the NER task, the format is
\begin{tiny}
\begin{minted}[frame=single, baselinestretch=1.2, 
highlightcolor=light-gray!40,
breaklines,
highlightlines={1,2,3,4,5,6}]{python3}
class NamedEntityRecognition:
    """ extract named entities from the input_text . """
    def __init__(self, input_text):
        self.input_text = "Steve became CEO of Apple in 1998 ."
        entity_list = []
        # extracted named entities
        entity_list.append({"text": "Steve", "type": "person"})
        entity_list.append({"text": "Apple",\ 
                        "type": "organization"})
\end{minted}
\end{tiny}

% \newpage
\noindent For the RE task, the format is
\begin{tiny}
\begin{minted}[frame=single, baselinestretch=1.2, 
highlightcolor=light-gray!40,
breaklines,
highlightlines={1,2,3,4,5,6}]{python3}
class RelationExtraction:
    """ extract the relations of named entities from the input_text . """
    def __init__(self, input_text):
        self.input_text = "Steve became CEO of Apple in 1998"
        entity_relation_list = []
        # extacted relations
        entity_relation_list.append({"rel_type": "work for", "ent1_type": "person", "ent1_text": "Steve", "ent2_type": "organization", "ent2_text": "Apple"})
\end{minted}
\end{tiny}

\noindent\texttt{3. func exec}: describe the IE tasks as a function execution procedure.\\

\noindent For the NER task, the format is
\begin{tiny}
\begin{minted}[frame=single, baselinestretch=1.2, 
highlightcolor=light-gray!40,
breaklines,
highlightlines={1,2,3,4}]{python3}
# extract named entities from a sentence .
input_text = "Steve became CEO of Apple in 1998 ."
output = named_entity_recognition(input_text)
# the output is
# {"text": "Steve", "type": "person"}
# {"text": "Apple", "type": "organization"}
\end{minted}
\end{tiny}

\noindent For the RE task, the format is
\begin{tiny}
\begin{minted}[frame=single, baselinestretch=1.2, 
highlightcolor=light-gray!40,
breaklines,
highlightlines={1,2,3,4}]{python3}
# extract the relations of named entities from from a sentence .
input_text = "Steve became CEO of Apple in 1998"
output = relation_extraction(input_text)
# the output is
# {"rel_type": "work for", "ent1_type": "person", "ent1_text": "Steve", "ent2_type": "organization", "ent2_text": "Apple"}
\end{minted}
\end{tiny}

\noindent\texttt{4. func init-}: perturb the rational format design by exchanging the format design of NER and RE tasks.\\

\noindent For the NER task, the format is
\begin{tiny}
\begin{minted}[frame=single, baselinestretch=1.2, 
highlightcolor=light-gray!40,
breaklines,
highlightlines={1,2,3,4,5}]{python3}
def relation_extraction(input_text):
    """ extract the relations of named entities from the input_text . """
    input_text = "Steve became CEO of Apple in 1998 ."
    entity_relation_list = []
    # extracted relations
    entity_relation_list.append({"text": "Steve", "type": "person"})
    entity_relation_list.append({"text": "Apple","type": "organization"})
\end{minted}
\end{tiny}

\newpage
\noindent For the RE task, the format is
\begin{tiny}
\begin{minted}[frame=single, baselinestretch=1.2, 
highlightcolor=light-gray!40,
breaklines,
highlightlines={1,2,3,4,5}]{python3}
def named_entity_recognition(input_text):
    """ extract named entities from the input_text . """
    input_text = "Steve became CEO of Apple in 1998"
    entity_list = []
    # extacted named entities
    entity_list.append({"rel_type": "work for", "ent1_type": "person", "ent1_text": "Steve", "ent2_type": "organization", "ent2_text": "Apple"})
\end{minted}
\end{tiny}

% \newpage
\subsection{Text Format Prompts}
\label{sec:text-prompts}

Similar to the above section (\ref{sec:code-prompts}), we describe the textual format prompt we used given the text input \textit{"Steve became CEO of Apple in 1998 ."}.
The text input prompts are all the same and we highlighted the expected outputs with {\color[RGB]{68,114,196}blue} colour.

~\\
\noindent\texttt{1. struct lang: } our mainly used text format prompt.

~\\
\noindent For the NER task, the transformed format is:
\begin{center}
\vspace{2pt}
\framebox{
\begin{minipage}{0.45\textwidth}
\fontsize{10pt}{12pt}\selectfont
\noindent The text is "Steve became CEO of Apple in 1998 .". The named entities in the text: \texttt{{\color[RGB]{68,114,196}((person: Steve)(organization: Apple))}}
\end{minipage}
}
\end{center}

~\\
\vspace{2pt}
\noindent For the RE task, the transformed format is:
\begin{center}
\vspace{2pt}
\framebox{
\begin{minipage}{0.45\textwidth}
\fontsize{10pt}{12pt}\selectfont
\noindent The text is "Steve became CEO of Apple in 1998 .". The relations of named entities in the text:  \texttt{{\color[RGB]{68,114,196}((person: Steve (work for: Apple)) (organization: Apple))}}
\end{minipage}
}
\end{center}

~\\

% \newpage
\noindent\texttt{2. natural lang: } a more "natural" format to describe the structures in natural language. 

~\\
\noindent For the NER task, the transformed format is:
\begin{center}
\vspace{2pt}
\framebox{
\begin{minipage}{0.45\textwidth}
\fontsize{10pt}{12pt}\selectfont
\noindent The text is "Steve became CEO of Apple in 1998 .". The named entities in the text: \texttt{{\color[RGB]{68,114,196}"Steve" is "person". "Apple" is "organization".}}
\end{minipage}
}
\end{center}
\vspace{2pt}

\noindent For the RE task, the transformed format is:
\begin{center}
\vspace{2pt}
\framebox{
\begin{minipage}{0.45\textwidth}
\fontsize{10pt}{12pt}\selectfont
\noindent The text is "Steve became CEO of Apple in 1998 .". The relations of named entities in the text: \texttt{{\color[RGB]{68,114,196}person "Steve" work for organization "Apple".}}
\end{minipage}
}
\end{center}

\begin{table*}
\centering
\resizebox{0.9\linewidth}{!}{
\begin{tabular}{lcccccccc}
\hline
\multicolumn{1}{c}{\multirow{2}{*}{\textbf{Model}}} & \multirow{2}{*}{\begin{tabular}[c]{@{}c@{}}\textbf{Prompt}\\ \textbf{Type}\end{tabular}} & \multicolumn{3}{c}{\textbf{Entity}}                    & \multicolumn{4}{c}{\textbf{Relation}}                                  \\ 
\cmidrule(l{3pt}r{3pt}){3-5}% \cline{3-9}
\cmidrule(l{3pt}r{3pt}){6-9}

% \cline{3-9} 
\multicolumn{1}{c}{}                       &                                                                        & \textbf{CoNLL03} & \textbf{ACE04} & \textbf{ACE05-E} & \textbf{CoNLL04} & \textbf{ACE05-R} & \textbf{NYT} & \textbf{SciERC} \\ \hline
$\text{UIE-large}$                                  & text                                            & $\text{46.75}_{\pm 6.13}$       & $\text{35.25}_{\pm 2.31}$     & $\text{34.29}_{\pm 1.93}$
& $\text{25.81}_{\pm 5.93 }$      & 
$\text{5.15}_{\pm 4.43}$           & -            & $\text{4.65}_{\pm 0.61}$    \\
Codex   & code                                  & $\textbf{75.89}_{\pm 1.79 }$     & $\textbf{54.27}_{\pm 2.14}$      & 
$\textbf{51.91}_{\pm 2.51}$          & $\textbf{43.05}_{\pm 3.42 }$        &           $\textbf{7.09}_{\pm 4.40 }$         & $\textbf{32.17}_{\pm 1.46}$  &
$\textbf{6.05}_{\pm 0.82}$
\\ \hline
\multicolumn{2}{c}{RoI$\uparrow$}   &
$\text{62.33}\; \%$     & 
$\text{53.96} \;\%$      & 
$\text{51.39}\; \%$          & 
$\text{66.80}\; \%$        &           
$\text{37.67} \;\%$         & 
-  &
$\text{30.11} \;\%$
\\ 
\hline
\end{tabular}
}
\caption{The 1-shot results of UIE-large and Codex, and Rate of Increase (RoI) of Codex than UIE-large.}
\label{tab:1-shot}
\end{table*}

\begin{table*}
\centering
\resizebox{\linewidth}{!}{
\begin{tabular}{lcccccccc}
\toprule
\multicolumn{1}{c}{\multirow{2}{*}{\textbf{Model}}} & \multirow{2}{*}{\begin{tabular}[c]{@{}c@{}}\textbf{Prompt}\\ \textbf{Type}\end{tabular}} & \multicolumn{3}{c}{\textbf{Entity}}                    & \multicolumn{4}{c}{\textbf{Relation}}                                  \\ 
\cmidrule(l{3pt}r{3pt}){3-5}% \cline{3-9}
\cmidrule(l{3pt}r{3pt}){6-9}
% \cline{3-9} 
\multicolumn{1}{c}{}                       &                                                                        & \textbf{CoNLL03} & \textbf{ACE04} & \textbf{ACE05-E} & \textbf{CoNLL04} & \textbf{ACE05-R} & \textbf{NYT} & \textbf{SciERC} \\ \hline
\texttt{text-davinci-002} & text   &   $\text{68.84}_{\pm 1.29}$ &   $\text{45.51}_{\pm 0.23}$  &   $\text{48.93}_{\pm 0.49}$    &  $\text{39.67}_{\pm 2.44}$         &   $\text{5.13}_{\pm 1.24}$       &        $\text{16.07}_{\pm 4.67}$                     &       $\text{4.39}_{\pm 0.98}$       \\
\texttt{code-davinci-002}                                      & code     & $\textbf{82.32}_{\pm 0.37}$          & $\textbf{55.29}_{\pm 0.37}$     &  $\textbf{54.82}_{\pm 2.09}$  
&  $\textbf{53.10}_{\pm 2.02}$    & $\textbf{14.02}_{\pm 3.27}$   & $\textbf{32.17}_{\pm 1.46}$    &    $\textbf{7.74}_{\pm 1.54}$                 \\ \hline
\texttt{text-davinci-001}                                  & text   & $\text{38.55}_{\pm \text{6.11}}$       & $\text{29.23}_{\pm \text{1.49}}$     & $\text{29.73}_{\pm \text{2.22}}$         & $\text{19.63}_{\pm \text{4.37}}$       & $\text{0.89}_{\pm \text{0.66}}$          & -            & $\text{0.87}_{\pm \text{0.22}}$       \\
\texttt{code-davinci-001}                              & code  & $\textbf{61.86}_{\pm \text{1.88}}$       & $\textbf{33.62}_{\pm \text{3.85}}$      & $\textbf{36.26}_{\pm \text{1.45}}$          & $\textbf{28.75}_{\pm \text{1.90}}$       &  $\textbf{1.65}_{\pm \text{1.55}}$ &    -     &    $\textbf{1.91}_{\pm \text{0.30}}$    \\ \hline
\end{tabular}
}
\caption{Performances of different LLMs. \texttt{text-davinci-001} is an InstructGPT model based on the previous GPT-3 model with Feedback Made Easy strategy. \texttt{code-davinci-001} is an earlier version of \texttt{code-davinci-002}.}
\label{tab:diff-elms}
\end{table*}

\begin{table*}
\centering
\resizebox{0.6\linewidth}{!}{
\begin{tabular}{lllllll}
\toprule
           & \multicolumn{3}{c}{\textbf{Entity Label Error}} & \multicolumn{3}{c}{\textbf{Entity Span Error}} \\
           \cmidrule(l{3pt}r{3pt}){2-4}
\cmidrule(l{3pt}r{3pt}){5-7}
           & CoNLL03     & ACE04    & ACE05-E    & CoNLL03     & ACE04     & ACE05-E    \\
           \midrule
\#test     & 3453        & 812      & 1060          & 3453        & 812       & 1060        \\ \midrule
\#in-context shot   & 5           & 2        & 2             & 5           & 2         & 2           \\
\midrule
GPT-3+text  & 15          & 298      & 414           & 113         & 140       & 114         \\
GPT-3+code  & 57          & 755      & 949           & 28          & 73        & 57          \\
Codex+text & 3           & 30       & 64            & 90          & 88        & 141         \\
Codex+code & 8           & 536      & 601           & 18          & 51        & 37          \\ \bottomrule
\end{tabular}
}
\caption{Detailed Errors on NER datasets. "Entity Label Error" means the predicted label is not in the predefined label set. "Entity Span Error" means the predicted span is not in the original input text. The reported error numbers are counted by summing 3 different seeds.}
\label{tab:ner-error}
\end{table*}

\begin{table*}
\centering
\resizebox{\linewidth}{!}{
\begin{tabular}{lllllllllllll}
\toprule
           & \multicolumn{4}{c}{\textbf{Ent1 Type Error}}                                    & \multicolumn{4}{c}{\textbf{Ent1 Span Error}}                                        & \multicolumn{4}{c}{\textbf{Relation Type Error}}                                                                                   \\
                      \cmidrule(l{3pt}r{3pt}){2-5}
\cmidrule(l{3pt}r{3pt}){6-9}
\cmidrule(l{3pt}r{3pt}){10-13}
           % \midrule
           & \multicolumn{1}{c}{CoNLL04} & \multicolumn{1}{c}{ACE05-Rel} & \multicolumn{1}{c}{NYT} & \multicolumn{1}{c}{SciERC} & \multicolumn{1}{c}{CoNLL04} & \multicolumn{1}{c}{ACE05-R} & \multicolumn{1}{c}{NYT} & \multicolumn{1}{c}{SciERC} & \multicolumn{1}{c}{CoNLL04} & \multicolumn{1}{c}{ACE05-R} & \multicolumn{1}{c}{NYT} & \multicolumn{1}{c}{SciERC} \\
           \midrule
\#test     & 288                         & 2050                           & 5000                    & 551                        & 288                         & 2050                           & 5000                    & 551                        & 288                         & 2050                           & 5000                    & 551                        \\
\midrule
\#in-context shot   & 5                           & 2                              & 1                       & 2                          & 5                           & 2                              & 1                       & 2                          & 5                           & 2                              & 1                       & 2                          \\
\midrule
GPT-3+text  & 2                           & 1078                           & 669                     & 335                        & 26                          & 266                            & 491                     & 160                        & 169                         & 617                            & 3274                    & 358                        \\
GPT-3+code  & 3                           & 410                            & 102                     & 410                        & 13                          & 105                            & 1029                    & 105                        & 6                           & 12                             & 2000                    & 50                         \\
Codex+text & 1                           & 815                            & 100                     & 815                        & 20                          & 155                            & 477                     & 155                        & 84                          & 820                            & 315                     & 820                        \\
Codex+code & 0                           & 346                            & 10                      & 346                        & 1                           & 108                            & 544                     & 108                        & 2                           & 0                              & 141                     & 17         \\
\bottomrule
\end{tabular}
}
\caption{Detailed Errors on RE datasets. "Ent1 Type Error" means the predicted entity type of the first entity is not in the predefined type set. "Ent1 Span Error" means the predicted span of the first entity is not in the original input text. "Relation Type Error" means the predicted label is not in the predefined relation type set. The reported error numbers are counted by summing 3 different seeds.}
\label{tab:re-error}
\end{table*}

\end{document}